\documentclass[a4paper,conference]{IEEEtran}
\usepackage{graphicx}
\usepackage{amsmath} 
\usepackage{booktabs}
\usepackage{multirow}
\usepackage{makecell}
\usepackage{array}
\usepackage{subcaption}
\usepackage{stfloats}
\usepackage{capt-of}
\usepackage{lipsum}
\usepackage{nidanfloat}

\usepackage{bbm}
\usepackage{bm}
\usepackage{color}
\usepackage{url}
\newsavebox{\bigimage}

\newcommand{\tabincell}[2]{\begin{tabular}{@{}#1@{}}#2\end{tabular}}
\ifCLASSINFOpdf
\else
\fi
\begin{document}
%

\title{Human Segmentation with Dynamic LiDAR Data}

 
\author{\IEEEauthorblockN{Tao Zhong\IEEEauthorrefmark{1},
Wonjik Kim\IEEEauthorrefmark{2}, Masayuki Tanaka\IEEEauthorrefmark{3} and
Masatoshi Okutomi\IEEEauthorrefmark{4}}
\IEEEauthorblockA{Department of Systems and Control Engineering, School of Engineering,\\
Tokyo Institute of Technology, Meguro-ku, Tokyo 152-8550, Japan\\
Email: \IEEEauthorrefmark{1}tzhong@ok.sc.e.titech.ac.jp,
\IEEEauthorrefmark{2}wkim@ok.sc.e.titech.ac.jp,
\IEEEauthorrefmark{3}mtanaka@sc.e.titech.ac.jp,
\IEEEauthorrefmark{4}mxo@sc.e.titech.ac.jp}}
\maketitle


%

\maketitle

\begin{abstract}

Consecutive LiDAR scans compose dynamic 3D sequences, which contain more abundant information than a single frame. Similar to the development history of image and video perception, dynamic 3D sequence perception starts to come into sight after inspiring research on static 3D data perception. This work proposes a spatio-temporal neural network for human segmentation with the dynamic LiDAR point clouds. It takes a sequence of depth images as input. It has a two-branch structure, i.e., the spatial segmentation branch and the temporal velocity estimation branch. The velocity estimation branch is designed to capture motion cues from the input sequence and then propagates them to the other branch. So that the segmentation branch segments humans according to both spatial and temporal features. These two branches are jointly learned on a generated dynamic point cloud dataset for human recognition. Our works fill in the blank of dynamic point cloud perception with the spherical representation of point cloud and achieves high accuracy. The experiments indicate that the introduction of temporal feature benefits the segmentation of dynamic point cloud. 

\end{abstract}


%
\IEEEpeerreviewmaketitle

\section{Introduction}
Accurate and fast 3D environment perception is of great importance in robotics field, as real-time geometric information of pedestrians is necessary for the following motion planning. Light Detection and Ranging (LiDAR) is especially attracting research and industry interest because of its large detection range, robustness under various conditions, descending price, etc. While early researches~\cite{guo20143d} used handcraft feature in point cloud processing, learning-based methods are also able to extract the features these days.

In most of the current datasets, the point cloud generated within one scan is formatted as one frame. Therefore, when the detection or segmentation is done frame by frame, each frame is treated as independent static 3D data. Many researchers have successfully applied deep learning methods to static point cloud perception. They~\cite{Guo2019DeepLF},~\cite{qi2017pointnet},~\cite{chen2017multi},~\cite{luo2018fast},~\cite{milioto2019rangenet++} extract spatially local and global features based on various representations of point cloud, such as point-wise method, voxel-based method and spherical representation method, etc. Subsequently, the features are used for classification, object detection, and semantic segmentation. 

\begin{figure}
\centering
\includegraphics[height = 5cm, width=\hsize]{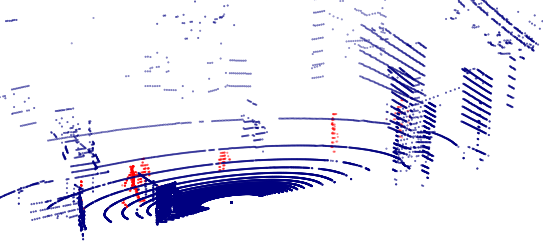}
\caption{An example of segmentation results on generated data. Blue points are classified as background, and red points are classified as human.}
\label{fig:fig001}
\end{figure}

In the real world, LiDAR rotates constantly, generating consecutive sequences of point cloud frames. These 3D videos are dynamic and thus contain abundant spatial and temporal cues. However, compared to the static point cloud, there is so far few researches regarding the dynamic point cloud. One of the reasons is the lack of annotated training data, since labeling millions of points in sequences costs undoubtedly a tremendous amount of time and resources. Recently, the release of several synthetic and real datasets~\cite{kim2020learning},~\cite{ros2016synthia},~\cite{behley2019semantickitti} for dynamic point cloud is accelerating some related researches. The point-wise method~\cite{liu2019meteornet} and voxel-based method~\cite{choy20194d}, which were already proved effective in static point cloud domain, have been extended to dynamic point cloud domain. They have demonstrated better performance than before, and showed the benefits of spatio-temporal features.

The proposed architecture in this work is inspired by the two-stream networks for unsupervised video object segmentation~\cite{cheng2017segflow}, which constructed communication of spatial and temporal features. With the help of a dataset that provides both per-point classification annotation and velocity annotation for synthetic LiDAR data, we train a network that can jointly predict human segmentation and velocity map in the dynamic point cloud. Consequently, different from the previous works that make use of the temporal information implicitly, we are able to extract it explicitly and leverage it in the segmentation task. Moreover, the proposed model is based on the spherical representation method, taking a sequence of 2D range images of point cloud sequences as input. To the best of our knowledge, our work is the first one that study the dynamic point cloud segmentation with this representation. 

The experiments show that the proposed model achieves higher accuracy than the existing state-of-the-art methods on the human segmentation dataset. 
The utilization of temporal cues effectively improves the performance of the human detection. We also investigate the effect of the length of frames which we use for human detection.

All data and sample code are available online\footnote{http://www.ok.sc.e.titech.ac.jp/res/LHD/}.

\section{Related Works}

\subsection{Static point cloud perception}

According to the way of handling the unordered point cloud, recent learning-based methods can be divided into two types: point-based networks~\cite{qi2017pointnet},~\cite{qi2017pointnet++},~\cite{hu2019randla} and projection-based networks~\cite{Guo2019DeepLF},~\cite{chen2017multi},~\cite{wu2018squeezeseg},~\cite{milioto2019rangenet++}. 

Point-based methods aim to directly process raw points. As the groundbreaking work, PointNet~\cite{qi2017pointnet} takes the coordinates of points as input. The global feature is obtained from the point-wise feature with a symmetric function. PointNet++~\cite{qi2017pointnet++} further takes the local structure into consideration by grouping spatially close points. The point-based method is good at extracting permutation-invariant features, but it also comes with relatively high computation and memory requirements.   

Projection-based methods first convert point cloud into intermediate representations. Projecting point cloud onto multi-view planes~\cite{chen2017multi} makes implementing 2D convolution on each view possible. View-wise features are then aggregated into the global feature. Voxelization of point cloud~\cite{maturana2015voxnet} generates 3D tensors, on which 3D convolution can be applied. Another widely used method is spherical representation~\cite{wu2018squeezeseg},~\cite{wu2019squeezesegv2}, ~\cite{wang2018pointseg}, \cite{milioto2019rangenet++}, which projects point cloud onto a spherical plane. The obtained depth image can then be fed into common convolutional networks. SqueezeSeg~\cite{wu2018squeezeseg} is a typical and efficient model of this kind. Regrettably, the temporal relationship between frames is ignored when they are considered separately. Based on spherical representation, our work captures the spatiotemporal cues in sequences of depth images.

\subsection{Dynamic point cloud perception}

Due to the difficulty of annotation, most of the dynamic point cloud datasets are generated with simulation. Kim et al. developed a pipeline for single frame~\cite{kim2019automatic}~\cite{kim2019automatic2} and sequence of point cloud data generation~\cite{kim2020learning}. SYNTHIA~\cite{ros2016synthia} dataset provided many road scene videos generated by a game engine. Behley et al.~\cite{behley2019semantickitti} annotated all the frames of KITTI's~\cite{geiger2013vision} odometry dataset and released the SemanticKITTI dataset.

The methods in static point cloud perception also apply to the dynamic condition with some adjustment. Meteornet~\cite{liu2019meteornet} is a point-wise method based on PointNet. They extended the spatial neighborhood to a temporal and spatial neighborhood when extracting the local features of points. The features were then used for downstream tasks: action recognition, segmentation, and scene flow estimation. PointRNN~\cite{fan2019pointrnn} applies recurrent models to aggregate local features from different time steps. As for the voxel-based method, Luo et al. ~\cite{luo2018fast} constructed 4D tensors by concatenating the occupancy grid of point cloud along the time axis. They then performed 3D convolution on the temporal dimension to capture motion features, which benefited the detection, tracking, and motion forecasting tasks. Choy et al.~\cite{choy20194d} restrained the computation cost of high-dimensional convolution by substituting the 2D convolution layers of UNet~\cite{ronneberger2015u} with sparse 4D convolution layers. However, when it comes to the accuracy of pedestrian segmentation, the improvement relative to the static method was not so significant. This indicates that extracting temporal features in such an implicit manner might fail. Compared with the above approaches, we leverage temporal information more explicitly.

\subsection{Video Object Segmentation}

In unsupervised video object segmentation task, there is no reference mask as guidance. Therefore, motion cues are necessary when searching for foreground objects. As a representative kind of motion field representation, optical flow~\cite{horn1981determining} is frequently used. Some researchers~\cite{nilsson2018semantic}, ~\cite{zhu2017deep} strengthened the feature by concatenating the warped feature map of previous frames to that of the current frame. Segflow~\cite{cheng2017segflow} constructed a two-stream network that predicts segmentation and optical flow simultaneously. They proved
that the bidirectional communication of feature maps between two streams can boost the performance of both tasks. Our method is inspired by this work, forcing the network to learn motion cues from the velocity estimation task.

\begin{figure*}
\centering
\includegraphics[width=\hsize]{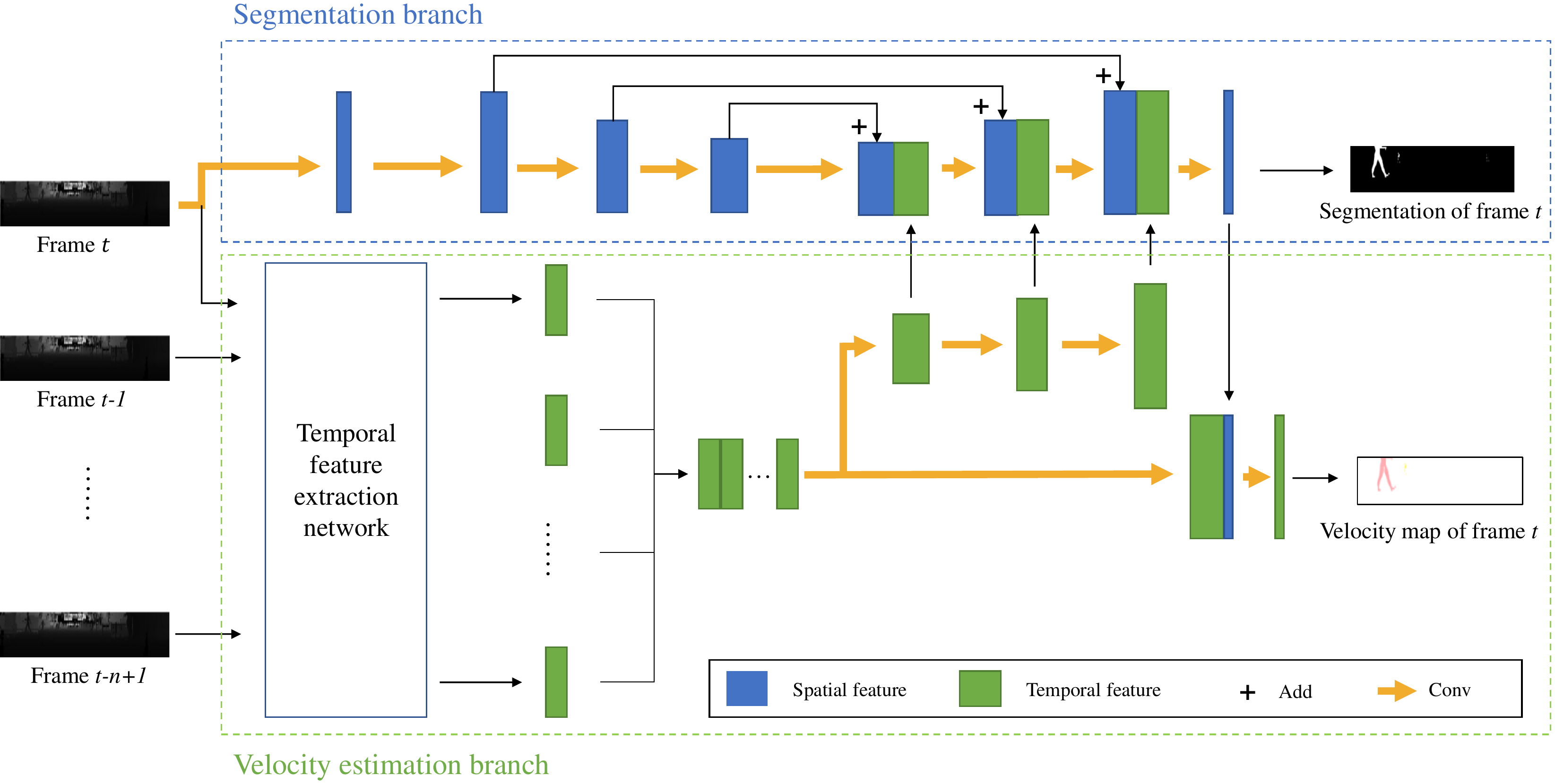}
\caption{The proposed network architecture which consists of the segmentation branch and the velocity estimation branch. They communicate spatial and temporal features in the expansion section. Frame $t$ denotes the current scan, while frame $t-1$ to $t-n+1$ denote the last $n-1$ scans.}
\label{fig:fig002}
\end{figure*}

\section{Proposed Method}

To segment humans in the dynamic point cloud with the help of motion information, we construct a two-branch network, one is the segmentation branch, and the other is the velocity estimation branch. It takes a sequence of depth images as input and predicts the class and velocity of each point in the last frame at one stage. In the following, we first explain the structure of each branch and the communication of temporal and spatial features between them. Then we briefly introduce how we generate sequential LiDAR data with segmentation and velocity annotation. The overall architecture of the proposed model is shown in Fig.~\ref{fig:fig002}.

\subsection{Segmentation Branch} 

The input of the segmentation branch is the last frame of the sequence. Therefore, it only extracts the spatial features and the size of the input tensor is $h$ $\times$ $w$ $\times$ $1$, where $h$ and $w$ are the height and width  of the range image. The output is the predicted probability distribution over classes for every pixel in this frame.

The segmentation branch has an hour-glass-shaped structure, consisting of a contraction section, an expansion section and skip connections. Considering that the resolution in the vertical direction is much lower than that in the horizontal direction, in the contraction section, the input frame is only downsampled in the horizontal direction. In our case, every two convolutional layers are followed by a max pooling layer with pool size [1 2], and the kernel size of convolution is $3$ $\times$ $3$. After three downsampling operations, the spatial feature maps with a size of $h$ $\times$ $\frac{1}{8}w$ $\times$ $C$ are upsampled back to the original image resolution in the expansion section. $C$ denotes the number of feature channels. In order to refine the segmentation result, the feature maps with size of $h$ $\times$ $w$ $\times$ $\frac{1}{8}C$, $h$ $\times$ $\frac{1}{2}w$ $\times$ $\frac{1}{4}C$, $h$ $\times$ $\frac{1}{4}w$ $\times$ $C$ before pooling layers in contraction section is added to those of the same size in the expansion section via skip connection. The channel number $C$ equals to 512. The activation function of the last layer is softmax.

\subsection{Velocity Estimation Branch}

The velocity estimation branch takes multiple consecutive frames as input, extracting temporal features. The size of the input tensor is $n$ $\times$ $h$ $\times$ $w$ $\times$ $1$, where $n$ represents the number of input frames. The output is the predicted velocity for every pixel in the last frame. 

Each input frame first goes through a weight shared feature extraction network. Its backbone is fully convolutional network~\cite{long2015fully}. The outputs are $n$ feature maps of corresponding input frames, the size of which is $h$ $\times$ $\frac{1}{16}w$ $\times$ $\frac{1}{8}C$. 

Since velocity is one kind of motion information, the feature maps in this branch are supposed to encode motion cues. Therefore, we concatenate them along the channel axis and regard the generated tensor as temporal feature maps. In order to make use of these features in the process of segmentation, we propagate the temporal feature maps to the segmentation branch. Specifically, they are up-sampled to match the size of spatial features in the expansion section and concatenated, illustrated by the green arrows in Fig.~\ref{fig:fig002}. In reverse, the predicted label map from the segmentation branch is also concatenated to the temporal feature maps as guidance. Finally, the last layer with linear activation decodes them into the $h$ $\times$ $w$ $\times$ $m$ velocity map after several convolutional layers. $m$ denotes the number of dimensions of the velocity vector, which in our case is two to represent the horizontal plane. 

\subsection{Loss Function}

For the loss function of the segmentation branch, we use the pixel-wise categorical cross-entropy.
For the loss function of the velocity estimation branch, we use Mean Square Error (MSE).
Hence, the total loss used in the optimization process is the combination of segmentation loss and velocity estimation loss for the background and human. The losses for background and human velocity estimation are computed separately. The defected pixels are not considered in the calculation.

Given $n$ input depth maps ${\bm d}_t$, ${\bm d}_{t-1}$, $\cdots$, estimated label map and velocity map for $t$-th frame can be expressed as
\begin{eqnarray}
 [\hat{\bm l}_t | \hat{\bm v}_t ] &=& {\bm f}( {\bm d}_{t}, \cdots, {\bm d}_{t-n+1} ; {\bm \theta} ) \,,
 \label{eq:yv}
\end{eqnarray}
where
$\hat{\bm l}_t$ is estimated label map for the $t$-frame,
$\hat{\bm v}_t$ is estimated velocity map for the $t$-frame,
and
${\bm f}( \cdots ;{\bm \theta})$ represents the network with weights ${\bm \theta}$. 
The total loss is defined as below:
\begin{eqnarray}
 L_t\! &=& \!
 \frac{\lambda_c}{N_c}
 \sum_{i} 
 {\rm CE}( \hat{l}_{t,i}, l_{t,i} )
 +
 \frac{\lambda_h}{N_h}
 \sum_{j=\{i|l_{t,i}=h\}}
 ||\hat{\bm v}_{t,j} - {\bm v}_{t,j}||_2^2
 \nonumber \\
 & &
 +
 \frac{\lambda_b}{N_b}
 \sum_{j=\{i|l_{t,i}=b\}}
 ||\hat{\bm v}_{t,j} - {\bm v}_{t,j}||_2^2
 \,, \label{eq:L_t}
\end{eqnarray}
where
${\rm CE}$ represents a cross-entropy,
$l_{t,i}$ is a true label for $i$-th pixel of $t$-th frame,
$||\cdot||_2^2$ represents L2 norm,
${\bm v}_{t,j}$ is a true velocity for $i$-th pixel of $t$-th frame,
$h$ and $b$ represent human and background labels,
$N_c$, $N_h$, and $N_b$ are the number of all pixels, human label pixels, and background pixels, 
and 
$\lambda_c$, $\lambda_h$, and $\lambda_b$ are hyper-parameters.

\subsection{Data Generation}

As an extension of 2D optical flow, scene flow~\cite{vedula1999three} denotes the translation vectors of points in 3D space. Unfortunately, as far as we know, there exists so far no point cloud dataset that provides both segmentation and scene flow annotation, due to the difficulty of labeling. Alternatively, we generate sequential LiDAR data with pixel-wise segmentation information and pixel-wise velocity information for training and evaluation, using the pipeline proposed in \cite{kim2020learning}. 

They are generated by combining the depth map of real background data and the depth map of synthetic human walking models~\cite{mochimaru2006dhaiba}. At the same time, the speed of LiDAR and human models is recorded. Consequently, each point is automatically labeled with its class and velocity. The points are classified into background and human. The velocity of a human point denotes the relative walking speed of the human model it belongs to, with respect to the origin of LiDAR at $x,y$ axis; the velocity of a background point is the relative speed of the background, i.e. $-v_{LiDAR}$. The forward direction of the LiDAR is designated as $x$ axis. 

\section{Experiments}

We evaluate the proposed network on a large-scale dataset for human segmentation. The performance is compared with multi-frame methods and a single-frame method. We also perform ablation studies on the influence of the velocity estimation branch and the length of the input sequence.

\subsection{Dataset and Training details}

The Automatic Labeled LiDAR Sequence dataset is composed of 1108 generated data sequences and 100 real data sequences. The background data of generated data is collected in the 3rd floor of Miraikan~\cite{Miraikan} using an HDL-32E LiDAR, and the real data is collected at the same place with pedestrians walking around. Each sequence is composed of 32 frames, the size of each frame is 32 $\times$ 1024. The data is accessible at  \url{http://www.ok.sc.e.titech.ac.jp/res/LHD/}

There are total 1108 generated sequences and 100 real sequences in our data set. We use 900 generated sequences for training, 100 generated sequences for validation. Then the networks are tested with the remaining 108 generated sequences and 100 real sequences. We train the network using Adam optimizer with an initial learning rate of 3e-5 and a decay rate of 3$\times$ $10^{-5}$. The batch size is set to 1 and 250 steps compose an epoch. The network is trained for 1000 epochs. The hyper-parameters $\lambda_c$, $\lambda_h$, $\lambda_b$ in Eq.~\ref{eq:L_t} are set as 1$\times$ $10^{5}$, 1, and 1001, in order to give more weights on the human points than background points. 

\subsection{Evaluation Metrics}

In consideration of the overwhelming number of background points, the performance is evaluated with intersection-over-union (IoU) score of human class, where points predicted as humans are regarded as positive. Hence, IoU is calculated with the total number of True Positives (TP), False Positives (FP), and False Negatives (FN) as below:
\begin{align}\label{eq2}
IoU = \frac{TP}{TP+FP+FN}    
\end{align}
Note that defected pixels are not considered for evaluation.

\subsection{Ablation Study}
We conduct ablation studies to verify the effect of components in the velocity estimation branch. Furthermore, the IoU score is calculated in several distance ranges in order to verify the effect at different distances. For instance, ’0 to 4 [m]’ means that only points 0 [m] and 4 [m] far from LiDAR are considered in the calculation.

\renewcommand{\arraystretch}{1.33}
\begin{table}[t]
\centering
\caption{Ablation study on components of velocity estimation branch. The performance is evaluated with IoU(\%) score of human class in different ranges.}
\label{table1}
\begin{tabular}{c|c|c|c|c|c}
\toprule[1pt]
\multicolumn{3}{c|}{Method} & \multicolumn{3}{c}{Proposed}    \\ \hline
\multicolumn{3}{c|}{Number of frames} & \multicolumn{3}{c}{4}  \\ \hline
\multicolumn{3}{c|}{Velocity estimation} &  & $\surd$ & $\surd$   \\ \hline
\multicolumn{3}{c|}{Temporal feature propagation} & $\surd$ &  & $\surd$   \\ \midrule[1pt]
\multirow{4}{*}{\makecell{Generated \\ data}} & \multirow{4}{*}{\makecell{Criteria\\range(m)}} & 0 to $\infty$ & 81.16
 & 77.11 & \textbf{86.08}  \\ \cline{3-6}
& & 0 to 4 & 92.21 & 90.98 & \textbf{94.06}   \\ 
& & 4 to 8 & 63.25 & 59.07 & \textbf{72.79}   \\ 
& & 8 to $\infty$ & 33.65 & 25.75 & \textbf{41.44}  \\
\hline
\multirow{4}{*}{\makecell{Real \\ data}} & \multirow{4}{*}{\makecell{Criteria\\range(m)}} & 0 to $\infty$ & 64.23 & 64.87 & \textbf{67.29}  \\ \cline{3-6}
& & 0 to 4 & 74.59 & 75.73 & \textbf{76.93}  \\
& & 4 to 8 & 49.84 & 51.83 & \textbf{53.79} \\ 
& & 8 to $\infty$ & 23.43 & \textbf{25.06} & 24.76   \\
\bottomrule[1pt]
\end{tabular}
\end{table}


When studying the effect of velocity estimation, we remove the last several layers for decoding the velocity map, i.e., the model only predicts segmentation result. When studying the effect of temporal feature propagation, we cut off the skip connections from the velocity estimation branch to the segmentation branch. As shown in Table~\ref{table1}, in the range of '0 to $\infty$ [m]', temporal feature propagation manages to bring about 9\% improvement on generated data and 2.4\% improvement on real data, respectively. Especially, in the '8 to $\infty$ [m]' range of generated data, velocity estimation and temporal feature propagation increase the IoU score by 7.9\% and 15.7\%, which indicates that motion cues help detection in the distance, where the point cloud is very sparse.

As shown in Fig. 5, the network can estimate the human segmentation and pixelwise velocity map. Then, we can derive the velocity map of the segmented area. The examples of human velocity estimation are illustrated in Fig. 8. According to the Fig. 8, the estimated velocities show similar tendency with ground truth.Asaconsequence,weconcludethatvelocityestimation with LiDAR only is a feasible task.

\renewcommand{\arraystretch}{1.33}
\begin{table*}[t]
\caption{Experiments on the effect of the number of input frames.}
\label{table2}
\centering
\def\svgwidth{\textwidth}
\begin{tabular}{c|c|c|c|c|c|c|c}
\toprule[1pt]
\multicolumn{3}{c|}{Method} & \multicolumn{5}{c}{Proposed}    \\ \hline
\multicolumn{3}{c|}{Number of frames} & 1 & 2 & 4 & 8 & 16  \\ \midrule[1pt]
\multirow{4}{*}{\makecell{Generated  data}} & \multirow{4}{*}{\makecell{Criteria range (m)}} & 0 to $\infty$ & 82.01 & 83.00 & 86.08 & \textbf{88.89} & 80.73\\ \cline{3-8}
& & 0 to 4 & 91.75 & 93.17 & 94.06 & \textbf{95.10} & 90.61 \\ 
& & 4 to 8 & 64.02 & 67.51 & 72.79 & \textbf{78.96} & 78.96 \\ 
& & 8 to $\infty$ & 38.05 & 34.52 & 41.44 & \textbf{46.52} & 31.96 \\
\hline
\multirow{4}{*}{\makecell{Real  data}} & \multirow{4}{*}{\makecell{Criteria range (m)}} & 0 to $\infty$ & 65.74 & 66.22 & 67.29 & \textbf{67.31} & 63.60 \\ \cline{3-8}
& & 0 to 4 & 75.85 & 77.20 & 76.93 & \textbf{77.43}  & 74.28\\ 
& & 4 to 8 & 52.50 & 52.77 & \textbf{53.79} & 51.53 & 46.01 \\ 
& & 8 to $\infty$ & 23.39 & 23.57 & \textbf{24.76} & 22.35 &23.61 \\
\bottomrule[1pt]
\end{tabular}
\end{table*}

\begin{figure*}[htbp]
\begin{subfigure}{\textwidth}
    \centering
    \includegraphics[width=\textwidth]{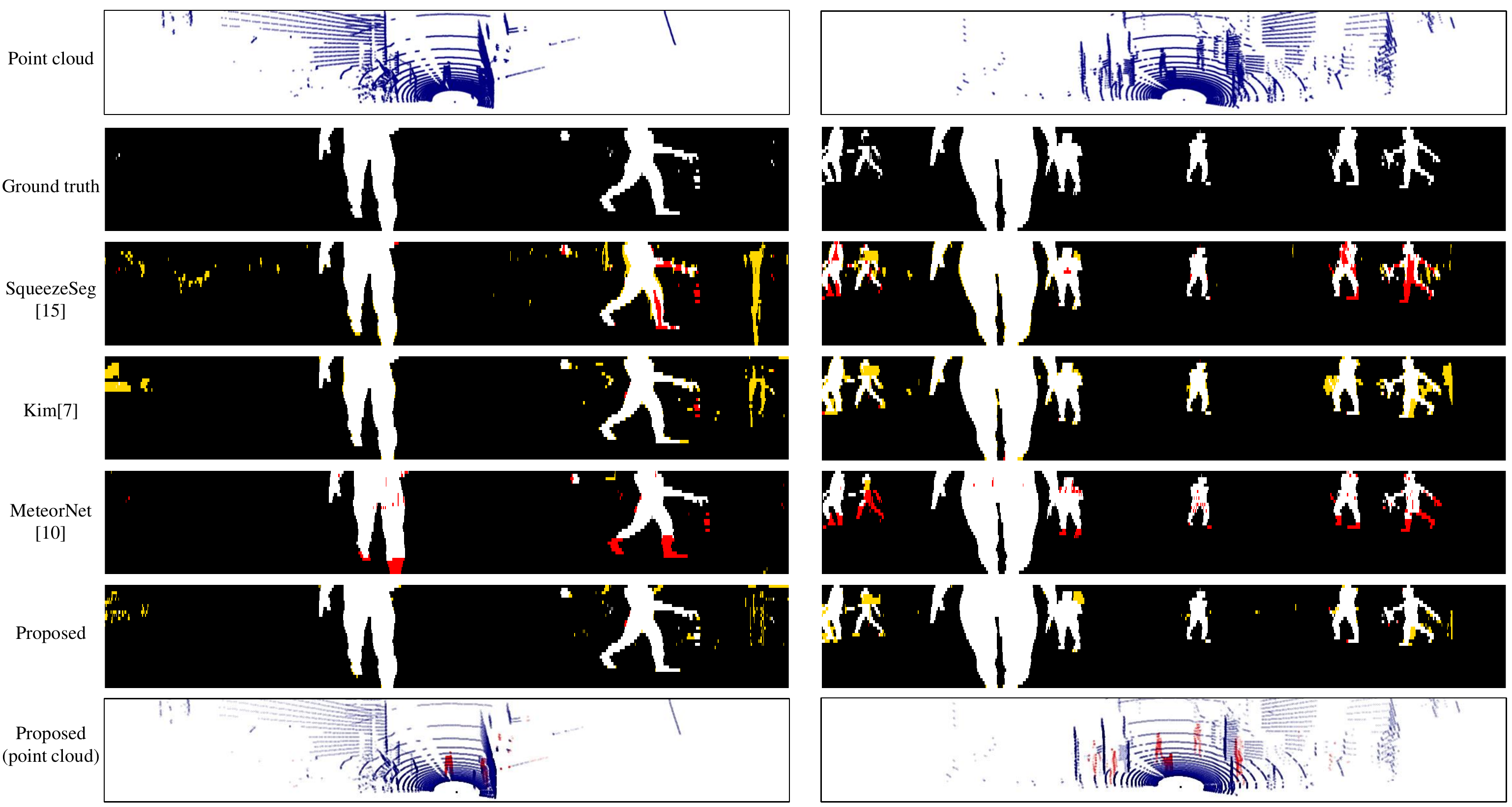}
    \caption{Generated data}
    \label{fig:sub-first}
\end{subfigure}
\vspace{0.8em}
\newline
\begin{subfigure}{\textwidth}
    \centering
    \includegraphics[width=\textwidth]{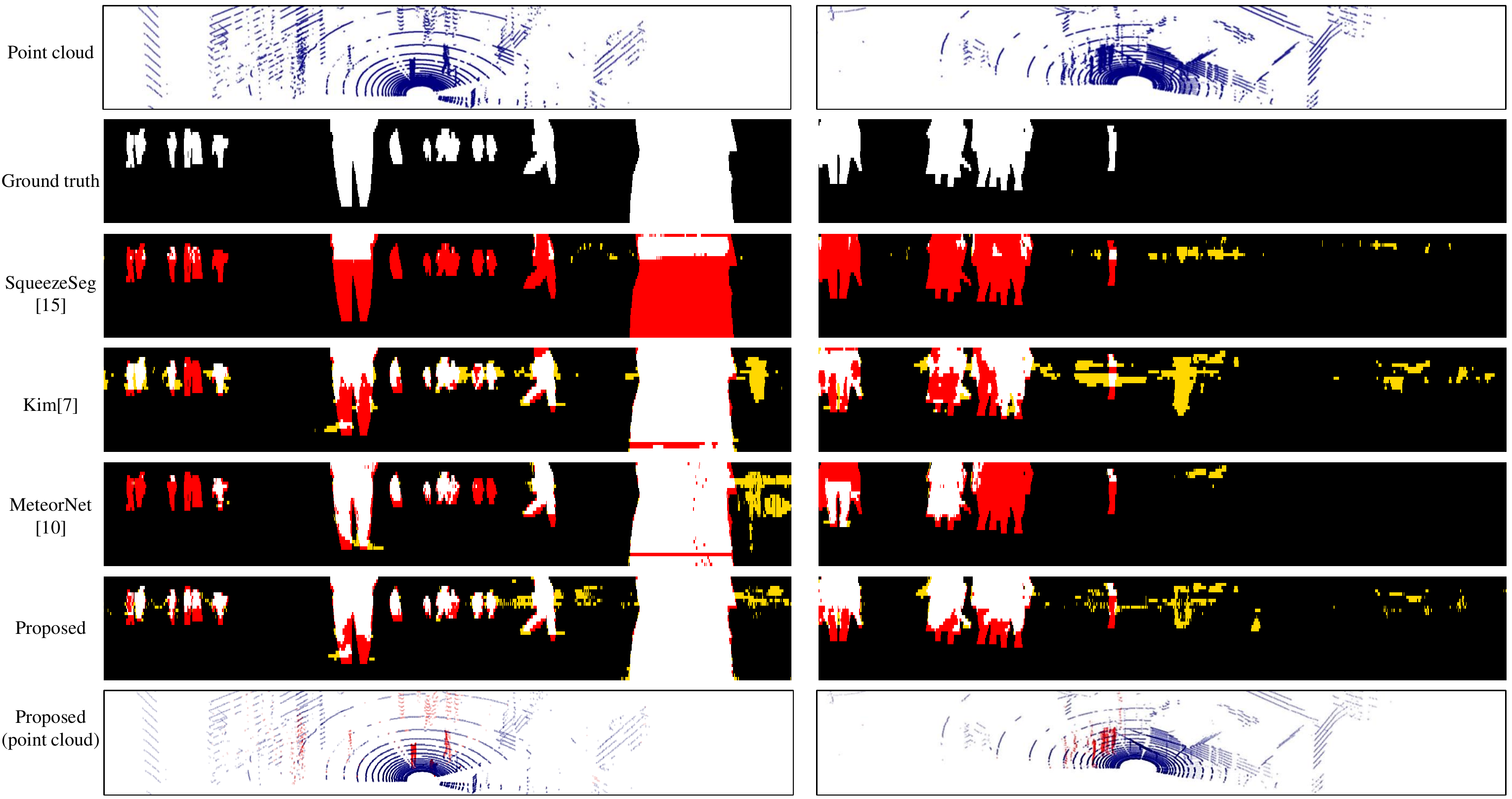}
    \caption{Real data}
    \label{fig:sub-second}
\end{subfigure}
\caption{Comparison of segmentation result on generated data and real data. In 2D plots, white area denotes True Positive and black area denotes True Negative, while red area denotes False Positive and yellow area denotes False Negative. }
\label{fig:fig003}
\end{figure*}

\subsection{Effect of the number of the input frames}
The relationship between the number of input frames and the performance is shown in Table~\ref{table2}. We evaluate the proposed model with 1-, 2-, 4-, 8-, 16- frame sequences as input. Note that there is no velocity estimation branch when the input is a single frame. 

When the length is not larger than eight frames, overall, accuracy increases along with the increase in the number of frames. On generated data, performance in all ranges is improved, and the IoU score in the range of '0 to $\infty$ [m]' with an 8-frame sequence input is 6.8\% higher than with a single frame input. This is attributed to the significantly better prediction accuracy in the range further than 4[m], which ulteriorly proves the advantage of temporal information in dynamic point cloud perception. This tendency is not so clear when it comes to real data. It might be caused by the domain gap between generated data and real data or inaccurate annotations in real data. However, an increase in number of frames also results in an increase in inference time and the accuracy drops significantly when the length reaches 16. Therefore, we consider 4 frame as an appropriate length.
\renewcommand{\arraystretch}{1.33}
\begin{table}[t]
\centering
\caption{Comparison of networks on the automatic labeled LiDAR sequence dataset for human segmentation. The performance is evaluated with IoU [\%] score of human class in the range of '0 to $\infty$ [m]'.}
\label{table3}
\begin{tabular}{c|c|c|c|c}
\toprule[1pt]
Method & {\makecell{SqueezeSeg\\(w/o CRF)~\cite{wu2018squeezeseg}}} & Kim~\cite{kim2020learning} & Meteornet~\cite{liu2019meteornet} & Proposed   \\ \hline
Input & depth & depth & xyz & depth   \\ \hline
\# of  frames & 1 & 4 & 4 & 4   \\ \midrule[1pt]
Generated  & 58.87 & 68.88 & 73.87 & \textbf{86.08} \\ 
Real & 11.35 & 58.69 & 52.20 & \textbf{67.29}  \\ \bottomrule[1pt]
\tabincell{c}{Run time \\ (ms)} & \textbf{6} & 46  & 840 & 51 \\ \bottomrule[1pt]
\end{tabular}
\end{table}
\subsection{Comparison}
The performance and runtime of the proposed network is compared to the model proposed by Kim and Meteornet whose input is a 4-frame sequence, as well as SqueezeSeg whose input is a single frame. We implement the SqueezeSeg without the CRF layer because it has better performance in human segmentation according to their paper. The results are shown in Table~\ref{table3}. Compared to the model from Kim, the proposed network improves the IoU score of the human class by nearly 17\% on generated data and by nearly 6\% on real data. It also accomplishes better accuracy than Meteornet and SqueezeSeg. This result shows the effectiveness of our model. Note that compared to the KITTI dataset used by SqueezeSeg and the SYNTHIA dataset used by Meteornet, our data contain neither intensity information nor color information. 
We used a single GTX 2080 Ti GPU and Intel Core i9 CPU. The run time of projection-based methods is shorter than the point-based method, Meteornet. For 4-frame input sequence, the proposed model runs at 51 ms/seq.

Figure~\ref{fig:fig003} is the 2D and 3D visualization of the segmentation results. Please refer to Figure~\ref{fig:fig001} for the 3D plot of prediction on generated data. It can be seen that the proposed model is able to recognize nearly all the human points of generated data with a few False Positives. On real data that are more difficult, the proposed model demonstrates its ability to restrain False Negative more obviously, which is important in the real robotics application.

\begin{figure}[htbp]
\begin{subfigure}{0.5\textwidth}
    \centering
    \includegraphics[height=1.2cm,width=\textwidth]{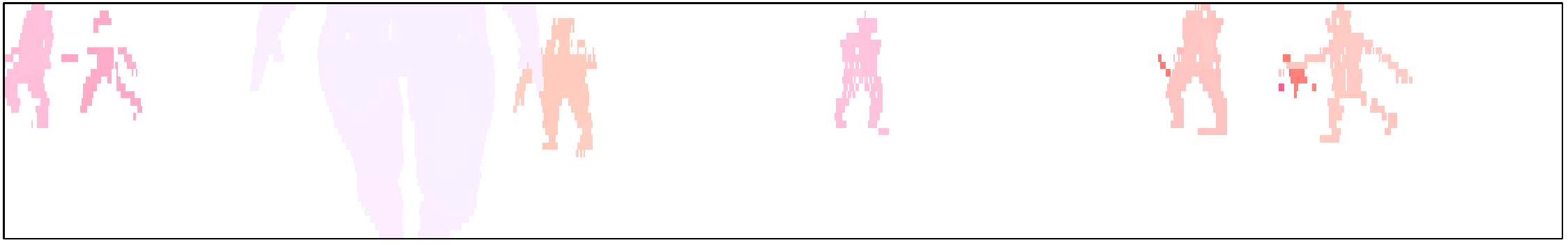}
    \caption{Ground truth}
    \label{fig:sub-first}
\end{subfigure}
\vspace{0.8em}
\newline
\begin{subfigure}{0.5\textwidth}
    \centering
    \includegraphics[height=1.2cm,width=\textwidth]{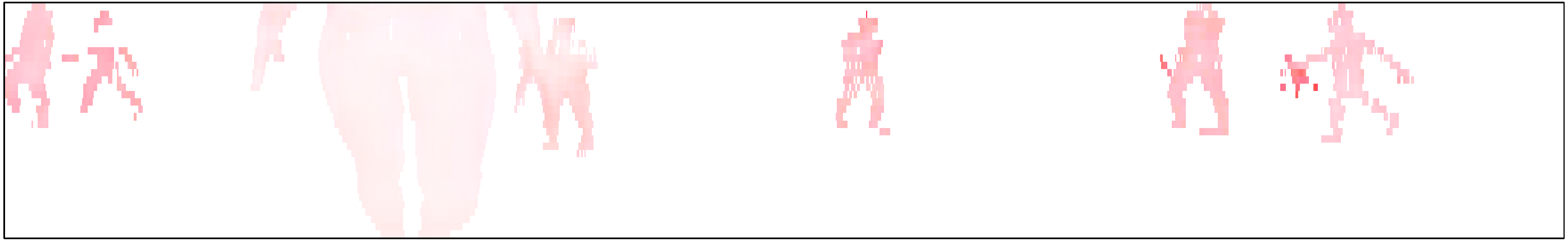}
    \caption{Predicted}
    \label{fig:sub-second}
\end{subfigure}
\newline
\begin{subfigure}{0.5\textwidth}
    \centering
    \includegraphics[width=0.3\textwidth]{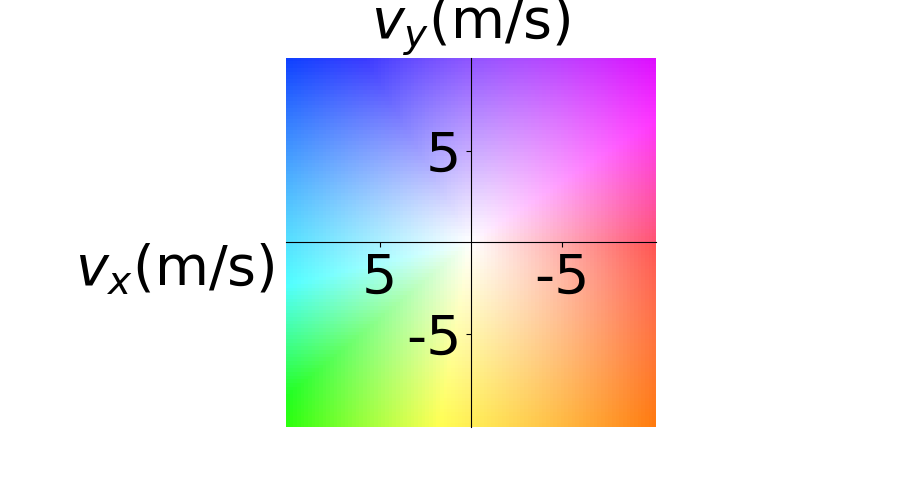}
    \caption{colormap}
    \label{fig:sub-second}
\end{subfigure}
\caption{Velocity map of human area.}
\label{fig:fig004}
\end{figure}

\subsection{Velocity estimation}
Our main goal is human segmentation, while the velocity map is also obtained as a by-product. The ground truth and prediction of velocity map of human area is shown in Figure~\ref{fig:fig004}, where the tendency of estimated velocities is similar to that of ground truth. This indicates that the velocity estimation branch captures motion information, which is used as cues for segmentation.

\section{Conclusion}
In this work, we have proposed a two-branch network for dynamic point cloud segmentation, which achieves high accuracy on a dataset for human segmentation. It is able to extract and utilize temporal information via feature propagations between the segmentation and the velocity estimation branches. According to the experimental results, we first conclude that temporal information contained in sequential data is beneficial to segmentation because motion cues can compensate for the sparseness of the input point cloud. Therefore, it is especially helpful for small and far object detection. Secondly, an appropriate increase in the length of sequence can improve the performance by producing more motion cues, while the trade-off between accuracy and computation cost exists.

For future improvement, we would consider how to improve the generalization ability of the model, so that it can be applied to more datasets. Also, we will try to reduce the computation load caused by the increase in the number of frames.

\bibliographystyle{IEEEtran}
\bibliography{ref}

%


\end{document}